\pdfoutput=1

\documentclass[11pt]{article}

\usepackage[final]{acl}

\usepackage{times}
\usepackage{latexsym}

\usepackage[T1]{fontenc}

\usepackage[utf8]{inputenc}

\usepackage{microtype}

\usepackage{inconsolata}

\usepackage{graphicx}
\usepackage{booktabs}
\usepackage{caption}
\usepackage{url}
\usepackage{hyperref}
\usepackage{microtype}
\usepackage{multirow}
\usepackage{wrapfig}
\usepackage{float}
\usepackage{afterpage} 
\usepackage{calc}
\usepackage{changepage}
\usepackage{capt-of}
\usepackage{dblfloatfix}
\usepackage{amsmath}

\newcommand{\dataname}{GroundCocoa}

\title{GroundCocoa: A Benchmark for Evaluating Compositional \& Conditional Reasoning in Language Models}

\author{Harsh Kohli \\
  \texttt{kohli.120@osu.edu} \\\And
  Sachin Kumar \\
  \texttt{kumar.1145@osu.edu} \\\And
  Huan Sun \\
  \texttt{sun.397@osu.edu} \\}

\begin{document}
\maketitle
\begin{abstract}
The rapid progress of large language models (LLMs) has seen them excel and frequently surpass human performance on standard benchmarks. This has enabled many downstream applications, such as LLM agents, to rely on their reasoning to address complex task requirements. However, LLMs are known to unexpectedly falter in simple tasks and under seemingly straightforward circumstances - underscoring the need for better and more diverse evaluation setups to measure their true capabilities. To this end, we choose to study compositional and conditional reasoning, two aspects that are central to human cognition, and introduce \dataname\ - a lexically diverse benchmark connecting these reasoning skills to the real-world problem of flight booking. Our task involves aligning detailed user preferences with available flight options presented in a multiple-choice format. Results indicate a significant disparity in performance among current state-of-the-art LLMs with even the best performing model, GPT-4 Turbo, not exceeding 67\% accuracy despite advanced prompting techniques.

\end{abstract}

\section{Introduction}


Conditional and compositional reasoning are central to navigating and interacting with complex systems through decision-making processes \citep{Oaksford2010CognitionAC, Simon1971HumanPS}. Conditional reasoning refers to the understanding and application of logical rules, often structured in ``if-then'' forms, which are fundamental to evaluating potential scenarios and anticipating outcomes in daily decision-making. Compositional reasoning involves solving complex problems by integrating solutions to simpler sub-problems in a structured manner. This cognitive process is crucial for understanding the relationships between different components of a task. We evaluate how effectively current LLMs exhibit these cognitive abilities, which are essential for both human and artificial intelligence. To that end, we introduce \textbf{\dataname}\footnote{\url{https://osu-nlp-group.github.io/GroundCocoa/}}, a benchmark designed to assess compositional and conditional reasoning within a grounding task.

\begin{figure*}[t]
    \centering
    \includegraphics[width=\textwidth]{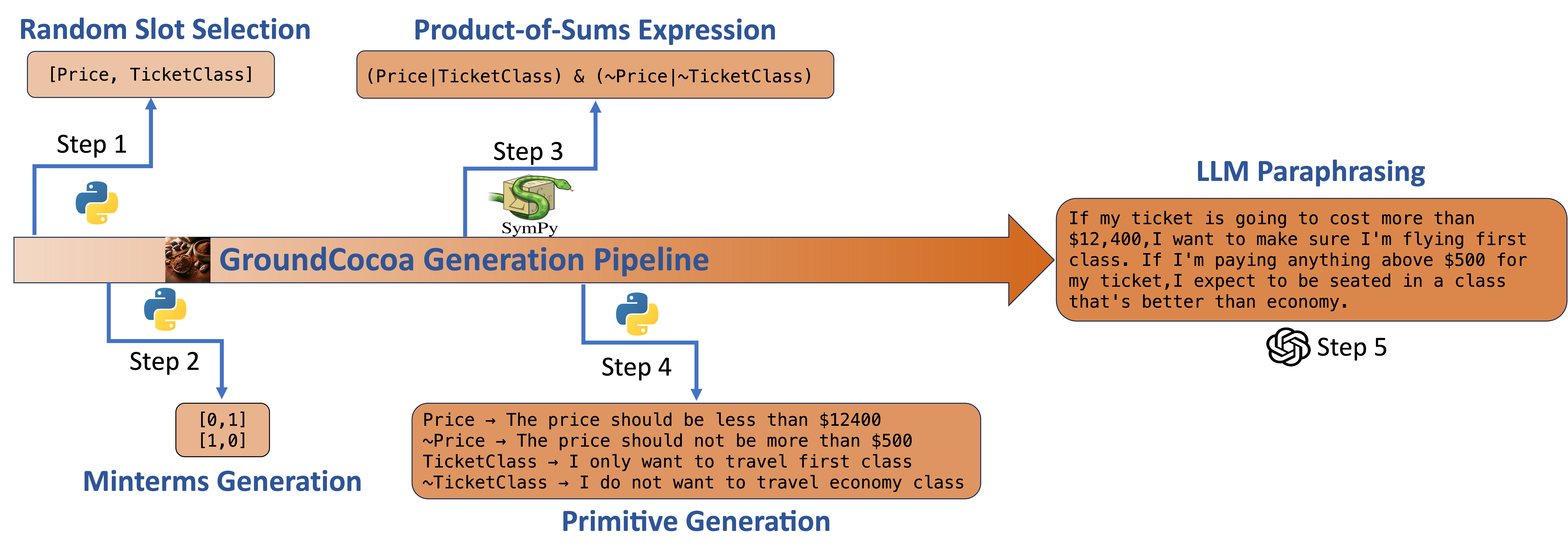}
    \caption{Stepwise depiction of \dataname\ query generation using 2 slots and 2 minterms.}
    \label{fig:cocoa_generation}
\end{figure*}

Set within a real-world inspired flight reservation scenario, \dataname\ comprises questions framed as user needs. Finding and booking flights is a complex task where user requirements might be many and highly convoluted. While we use flight booking to illustrate our idea, the compositional primitives forming our user requirements test for general skills such as temporal reasoning (e.g., "I want a flight departing after 5 pm") and mathematical reasoning (e.g., "Ticket price should be under \$1000"). Thus, we posit that results and insights derived from evaluating LLMs on \dataname\ should largely be applicable to other domains.

We leverage a controllable method, illustrated in Figure~\ref{fig:cocoa_generation}, to create samples of varying complexity. Our data generation process (\S\ref{sec:data_creation})
consists of a 5-stage pipeline including online scraping, constraint generation, symbolic logic to impose conditionality, paraphrasing user requirements, and matching generated requirements to available flight options. To test for robustness, we allow requirements to freely condition on one another and impose no restrictions on their nature. Additionally, we isolate a subset of more atypical queries that contain unconventional user needs (e.g., ``I want at least 2 layovers'') and evaluate their impact on model performance.

In addition to the release of the dataset and accompanying results, our contribution also includes the data generation pipeline which can be used to controllably generate samples of increasing complexity to challenge more advanced models in the future. Through slight modifications to the data scrapers and the primitive rule-set (described in Section~\ref{sec:data_creation}), the method can also be extended to incorporate other domains for a more diverse evaluation setup. Statistics of \dataname\ are shown in Table~\ref{tab:dataset_statistics}. Our key findings are as follows:

\begin{enumerate}
  \item Accuracy among contemporary LLMs varies greatly, ranging from a little better than random guess to about 67\% on a five-option multiple-choice question task. Within this spectrum, GPT-4 Turbo \citep{OpenAI_GPT4_2023} stands out, demonstrating a superior capacity of the GPT line of models to adapt and excel in novel reasoning tasks. However, conditional reasoning poses a significant challenge to all evaluated models, even on samples of relatively lower complexity. 
  \item Incorporating prompting techniques such as Chain of Thought (COT) \citep{wei2022chain} and Least-to-Most (L2M) prompting \citep{zhouleast} leads to mixed results, with only a modest performance improvement in some cases. Prior research has noted that although these methods help decompose problems into steps, LLMs struggle as the complexity of the individual steps grows \citep{hendrycksmath2021, madaan2022text, nogueira2021investigating, inproceedingslim}. These assertions hold true in our observations.
\item Including unconventional user requirements leads to a drop in accuracy of as much as 6\% in GPT-4 Turbo, indicating a training bias towards more typical needs.

\end{enumerate}

\section{Approach}
\label{sec:data_creation}

Figure~\ref{fig:cocoa_generation} illustrates our proposed approach for generating a user requirement. The task involves matching this generated requirement against 5 flight options where only 1 of the options satisfies the generated criteria. Our 5-stage data creation pipeline is detailed in subsequent sections. In the process of generating a natural language user requirement for flight booking, we are faced with the following considerations:  

\noindent \textbf{Conditionality of Constraints.} We aim to challenge contemporary models in their ability to reason through scenarios characterized by conditional complexity. This is done through mutual dependence of flight attributes which we refer to as \textit{slots}. As illustrated in the final requirement (Step 5) of Figure~\ref{fig:cocoa_generation}, there is an interdependence between the values for price and ticket class. This is a direct result of the generated minterm table. A minterm is a specific type of logical expression that represents exactly one row in a truth table where the function evaluates to true (1). In the context of GroundCocoa, each minterm represents a specific combination of flight attributes, or 'slots', that satisfies a particular user requirement. We represent this interdependence in logical form through a Product-of-Sums (POS) expression which consists of multiple \textsf{OR} operations (sums) which are later combined through \textsf{AND} operations (products). This process is further explained in a subsequent section (\S\ref{subsec:pos_gen}). The inclusion of \textsf{OR} operations between slots introduces conditional complexity to our user requirement, necessitating consideration of potential slot values in if-then scenarios. On the other hand, a greater number of \textsf{AND} conditions implies a higher number of variables that a model has to simultaneously reason over resulting in increased compositional complexity. 

\noindent\textbf{Satisfiability of POS Expression.} While generating the logical form for a user requirement, we must ensure satisfiability of the generated POS expression. For this, we use SymPy \citep{10.7717/peerj-cs.103}, an open-source Python symbolic mathematics library which generates an optimal POS expression given a minterm table (\S\ref{subsec:pos_gen}).

\noindent\textbf{Fuzziness in Slot Values.} Corresponding to each occurrence of a slot in the POS expression there has to be a unique constraint. For the example in Figure~\ref{fig:cocoa_generation}, the two constraints on the \textsf{price} slot are \{$<$$12400$, $<$$500$\}. We impose these constraints randomly through specialized rule-based systems corresponding to each slot. However, these might cause the final user criteria to become impossible to satisfy even if the corresponding POS expression is satisfiable. Thus, for a generated user requirement we perform checks to ensure that there exists at least 1 route that satisfies the criteria and at least 4 that do not so that there are at least 1 positive and 4 negative options for a generated requirement.

In addition to the test set, we also include a separate validation set which may be used for tuning hyperparameters. The pipeline may be reused to generate more complex samples in the future, and could also be extended to other domains through a slight modification of the data collection (\S\ref{subsec:data_collection}) and primitive generation (\S\ref{sec:primitive_generation}) stages.

\subsection{Flight Data Collection}
\label{subsec:data_collection}

We start with a list of the top 50 busiest airports by passenger traffic derived from \href{https://en.wikipedia.org/wiki/List\_of\_busiest\_airports\_by\_passenger\_traffic}{Wikipedia}. We choose source and destination airports  randomly from this list. A fixed departure date is also chosen randomly from the future and set for each flight search. The source, destination, and travel date are input to \href{https://www.google.com/travel/flights}{Google Flights}. We then sample a small number of flights from the search results. The sampled flights are chosen from each of economy, business, and first class and, for each flight option, all the relevant details such as the number of layovers, price, departure and arrival times etc. are saved. A sample flight schema with all the elements is provided in Appendix \ref{sec:schema}. We use \href{https://www.selenium.dev/documentation/webdriver/}{Selenium Webdriver} for scraping this data.

\subsection{Product-of-Sums Generation}
\label{subsec:pos_gen}

To generate a POS expression, we first randomly select a small number of flight attributes or \textit{slots}. The complete set of slots $S$ is as follows:

$S$=\{\textsf{airline}, \textsf{ticket class}, \textsf{departure time}, \textsf{arrival time}, \textsf{total travel time}, \textsf{number of layovers}, \textsf{average carbon emission difference}, \textsf{travel date}, \textsf{price}, \textsf{layover locations}, \textsf{layover times}\}

We vary the number of slots between 2 and 6 in order to generate samples of differing complexity. We then randomly generate 2-3 ``minterms'', the list of all input combinations of slots that generate a true (1). A higher number of minterms results in a greater conditional complexity. 
The slot symbols and generated minterms are input to SymPy which uses a redundant-group eliminating algorithm to output the smallest POS expression consistent with the minterm table.

\begin{table*}[t]
    \centering
    \resizebox{0.7\textwidth}{!}{ 
    \begin{tabular}{lccccccc}
    \toprule
         \multirow{2}*{Statistic} &  \multicolumn{6}{c}{(slot, minterm) configurations}& \multirow{2}*{\textbf{Total}}\\
         \cmidrule{2-7}
         &  (2,2)&  (3,2)&  (4,2)&  (4, 3)&  (5,2)&  (6,2)& \\
         \cmidrule{1-8}
         Test Samples&  1511&  1083&  710&  723&  451&  371& \textbf{4849}\\
         Test Unique Queries&  124&  136&  117&  129&  121&  101& \textbf{728}\\
         Val. Samples&  17&  17&  8&  5&  2&  3& \textbf{52}\\
         Val. Unique Queries&  1&  1&  1&  1&  1&  1& \textbf{6}\\
         Avg. Query Length& 65.04& 88.33& 103.88& 119.14& 124.56& 148.87&\textbf{95.95}\\
         Avg. Context Length&  -&  -&  -&  -&  -&  -& \textbf{1252.27}\\
         Vocab Size&  -&  -&  -&  -&  -&  -& \textbf{4200}\\
    \bottomrule
    \end{tabular}}
    \caption{Key Statistics of GroundCocoa.}
    \label{tab:dataset_statistics}
\end{table*}

\subsection{Primitive Generation}
\label{sec:primitive_generation}

Corresponding to each slot, we have developed a rule-based system that randomly imposes constraints on its values. These constraints are converted to natural language through templates. Since a POS expression may contain a negation, we generate two primitives at each turn - one for the constraint and one for its negation. A sample primitive for total travel time is shown in Table~\ref{tab:sample_constraint}.

\begin{table}[H]
    \small
    \centering
    \resizebox{\columnwidth}{!}{ 
    \begin{tabular}{lc}
    \toprule
         \textsf{TravelTime} & Travel Time should be more than 22 hours and 30 minutes.\\ 
    \midrule
         $\neg$\textsf{TravelTime} & Travel Time should not be more than 22 hours and 30 minutes.\\ 
    \bottomrule
    \end{tabular}}
    \caption{Sample primitive for total travel time.}
    \label{tab:sample_constraint}
\end{table}

At this stage, we also isolate samples that include any one of the following three primitives - (1) carbon emissions must be above the average for that route, (2) price of the flight must be above a minimum threshold, and (3) number of layovers on the route should be greater than a minimum. While this list is not exhaustive, such samples (henceforth referred to as "atypical" queries) are able to successfully encapsulate contrarian needs that are unlikely to manifest often during pretraining.

\subsection{LLM Paraphrasing and Human Validation}

We paraphrase the user requirement derived from rule templates to make them more natural-sounding while preserving the original intent and meaning. LLM paraphrasing is carried out in two distinct steps described below. The exact prompts and an example of intermediate results are provided in Appendix \ref{sec:llm_paraphrase}. We manually verify each query to ensure it is consistent with the primitives and make changes wherever necessary. 

\begin{enumerate}
  \item Individual primitives are substituted into each \textsf{sum} term and combined using templated rules. We then use GPT-4 Turbo to paraphrase each of the sum terms.
  \item  Next, we combine the individual sum terms into a \textsf{product} (logical \textsf{AND}). This is done by merging the paraphrases of sum terms, separated by periods. The resulting flight requirement is again paraphrased with GPT-4 Turbo.
\end{enumerate}

\subsection{Option Matching}
\label{op_matching}

We match the generated user requirements with the flight data collected in Section~\ref{subsec:data_collection}. Each route between the source and destination represents a potential choice in our multiple-choice dataset. Choices are divided into subsets containing one positive (matching the user requirement) and four negative (not matching the user requirement) options. This is done to ensure that each multiple-choice question has only a single correct answer for ease of evaluation. Many such subsets may be created from a single user requirement and, consequently, our dataset consists of queries repeated multiple times with differing choices. Table~\ref{tab:dataset_statistics} contains details of the number of unique queries and overall samples corresponding to all the slot/minterm configurations in \dataname.

\section{Results}

To measure performance on \dataname, we test several models of different sizes including both open-source and closed-source LLMs - LLAMA 2-chat \citep{touvron2023llama} / LLAMA 3-Instruct \citep{dubey2024llama}, Mixtral 8x7B - Instruct \citep{jiang2024mixtral} / Mistral 7B Instruct \citep{jiang2023mistral}, Gemini Pro \citep{team2023gemini}, and GPT-4 Turbo. Results from our experiments are shown in Table~\ref{tab:results_main}. We have 3 different evaluation setups for the our models - direct prompting, chain-of-thought (CoT) \citep{wei2022chain} prompting, and least-to-most (L2M) prompting \cite{zhouleast}. We aim to evaluate the intrinsic reasoning ability of current LLMs and, thus, exclude methods such as Program of Thoughts \cite{chen2022program} and Program-aided Language Models \cite{10.5555/3618408.3618843} that offload the critical reasoning component to an external engine such as a Python interpreter.

\subsection{Direct Prompting}

The models are presented with a sample from \dataname\ consisting of  a user requirement and 5 flight options in a zero-shot manner. The smaller models in our experiments are only evaluated in this setting. 

\subsection{Chain-of-Thought Prompting}

Since our task involves grounding user requirements to each answer choice, the CoT explanations are provided for each flight option given the user requirement. Thus, our standard CoT (CoT-full) consists of 5 distinct explanations. On GPT-4, we empirically observe that the large resulting context length can prove detrimental to model performance with the models often confusing between the requirements and options of the test case and the exemplar. To address this, we try a different prompting strategy (CoT-partial) with only two flight choices (1 positive and 1 negative) for the in-context example. Due to limitations on context length (4096 tokens) we are unable to run LLAMA 2-chat 70B on CoT-full. The exact prompts are given in Appendix \ref{sec:testing_prompts}. Results from our experiments are shown in Table~\ref{tab:results_main}. As alluded to previously, \dataname\ poses a significant challenge for each of the evaluated models, even with CoT prompting. The CoT-partial strategy leads to better results than CoT-full in 3 out of 4 cases, and best results are obtained using GPT-4 Turbo with CoT-partial. It is noteworthy, though, that there exists a marked difference in performance between competing models. Such variation represents a significant departure from the usual performance patterns observed in popular benchmarks such as MMLU \citep{hendrycks2021measuring}, HellaSwag \citep{zellers2019hellaswag}, ARC Reasoning Challenge \citep{clark2018think}, WinoGrande \citep{10.1145/3474381}, and GSM-8K \citep{cobbe2021gsm8k} among others, where results are much more comparable.

\subsection{Least-to-Most Prompting}

Finally, we do a limited evaluation with least-to-most prompting which carries out task decomposition through an iterative prompting procedure. The problem (user requirement) is broken down into multiple sub-problems and each sub-problem is solved iteratively through successive prompts. The number of decomposition steps (turns) required in L2M scales linearly with the compositional complexity of each sample. The large number of turns per sample leads to a higher inference cost. We thus test each of our larger models using L2M using a subset of 200 samples from \dataname\ - the corresponding rows are marked with an asterisk ($^{*}$) in Table~\ref{tab:results_main}. Results indicate that \dataname\ remains a challenging benchmark despite such multi-turn prompting methods for problem decomposition.

\begin{table}[h]
    \small
    \centering
    \resizebox{\columnwidth}{!}{ 
    \begin{tabular}{lccc}
    \toprule
         \multirow{1}*{} &  \multicolumn{1}{c}{Regular} & \multicolumn{1}{c}{Atypical} & \multirow{1}*{Total}\\
         \midrule
         Open-source Models & & & \\
         \cmidrule{1-1}
LLAMA 2-chat 7B & 14.56 & 14.66 & 14.60\\
LLAMA 3.1-chat 8B & 33.66 & 35.25 & 34.29\\
Mistral 7B Instruct & 25.70 & 26.10 & 25.86\\
LLAMA 2-chat 13B & 16.33 & 16.06 & 16.23\\
Mixtral 8x7B-Instruct & 45.79 & 42.48 & 44.48\\
Mixtral 8x7B-Instruct + CoT-full & 34.38 & 32.65 & 33.69\\
Mixtral 8x7B-Instruct + CoT-partial & 41.38 & 39.85 & 40.15\\
Mixtral 8x7B-Instruct + L2M$^{*}$ & 22.32 & 15.90 & 19.50\\
LLAMA 2-chat 70B & 24.13 & 21.63 & 23.13\\
LLAMA 2-chat 70B + CoT-partial & 25.73 & 23.97 & 25.03\\
LLAMA 3.1-chat 70B & 59.57 & 55.64 & 58.01\\
LLAMA 3.1-chat 70B + CoT-full & 58.37 & 57.67 & 58.09\\
LLAMA 3.1-chat 70B + CoT-partial & 60.22 & 58.66 & 59.60\\
LLAMA 3.1-chat 70B + L2M$^{*}$ & 68.18 & 50.89 & 58.50\\
         \midrule
         Closed-source Models & & & \\
         \cmidrule{1-1}
Gemini Pro & 42.79 & 40.46 & 41.86\\
Gemini Pro + CoT-full & 41.14 & 40.87 & 41.04\\
Gemini Pro + CoT-partial & 34.82 & 33.85 & 34.44\\
Gemini Pro + L2M$^{*}$ & 42.86 & 40.46 & 41.90\\
GPT-4 Turbo & 64.66 & 58.81 & 62.34\\
GPT-4 Turbo + CoT-full & 65.07 & 61.51 & 63.66\\
GPT-4 Turbo + CoT-partial & 67.77 & 65.62 & 66.92\\
GPT-4 Turbo + L2M$^{*}$ & 46.43 & 53.41 & 49.50\\
    \bottomrule
    \end{tabular}}
    \caption{Accuracy (\%) on \dataname.}
    \label{tab:results_main}
\end{table}

\section{Analysis}

Beyond assessing the overall model performance, we also investigate the consequences of varying the complexity of user criteria and presenting relatively unconventional user needs.

\subsection{Impact of Increasing Complexity}

In our analysis, we observe the performance of GPT-4 Turbo, the best-performing model from among those tested on \dataname\, across different levels of conditional and compositional complexity. In their recent work on assessing the limitations of transformer on compositional tasks, \citet{dziri2023faith} use computational graphs as approximations of the underlying reasoning processes in such models. They define the terms \textit{reasoning depth}, the length of the deepest layer in the computational graph from the source nodes, and \textit{reasoning width}, the mode of number of nodes in each layer - indicating the extent of multi-hop reasoning and compositional parallelism required to solve a given problem. Considering the characteristics of \dataname\, we focus on reasoning width - the number of variables a model has to simultaneously reason over for a given problem. Intuitively, this may be represented by the number of \textit{slots} used during the generation of a particular sample as described in Section~\ref{subsec:pos_gen}. However, keeping the number of rows in the minterm table constant while increasing the slots may often lead to lower conditional complexity as the number of slots is increased.

\begin{figure}[h]
    \centering
    \includegraphics[width=\columnwidth]{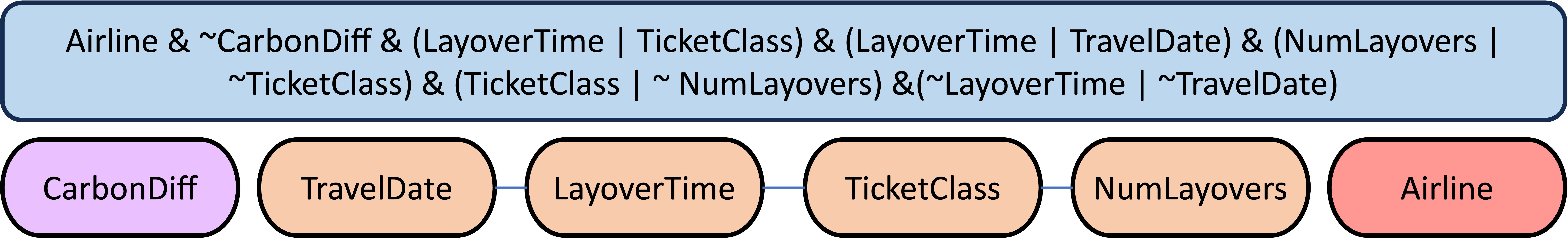}
    \caption{POS expression and its dependency graph.}
    \label{fig:dependency}
\end{figure}

In order to effectively gauge the compositional and conditional complexity of a sample in our dataset, we define a dependency graph derived from the POS expression corresponding to that sample. Vertices represent slots and a dependency (edge) is created when a particular slot co-occurs with another slot within a sum term in the POS. A sample POS expression and its corresponding dependency graph are shown in Figure~\ref{fig:dependency}. The graph has 3 connected components with the largest connected component (LCC) of size 4. The maximum degree is 2 which corresponds to the two connections for nodes \textsf{LayoverTime} and \textsf{TicketClass}.

\begin{figure}[h] 
  \centering 
  
  \begin{minipage}{0.35\columnwidth} 
    \centering
    \includegraphics[height=2.6cm]{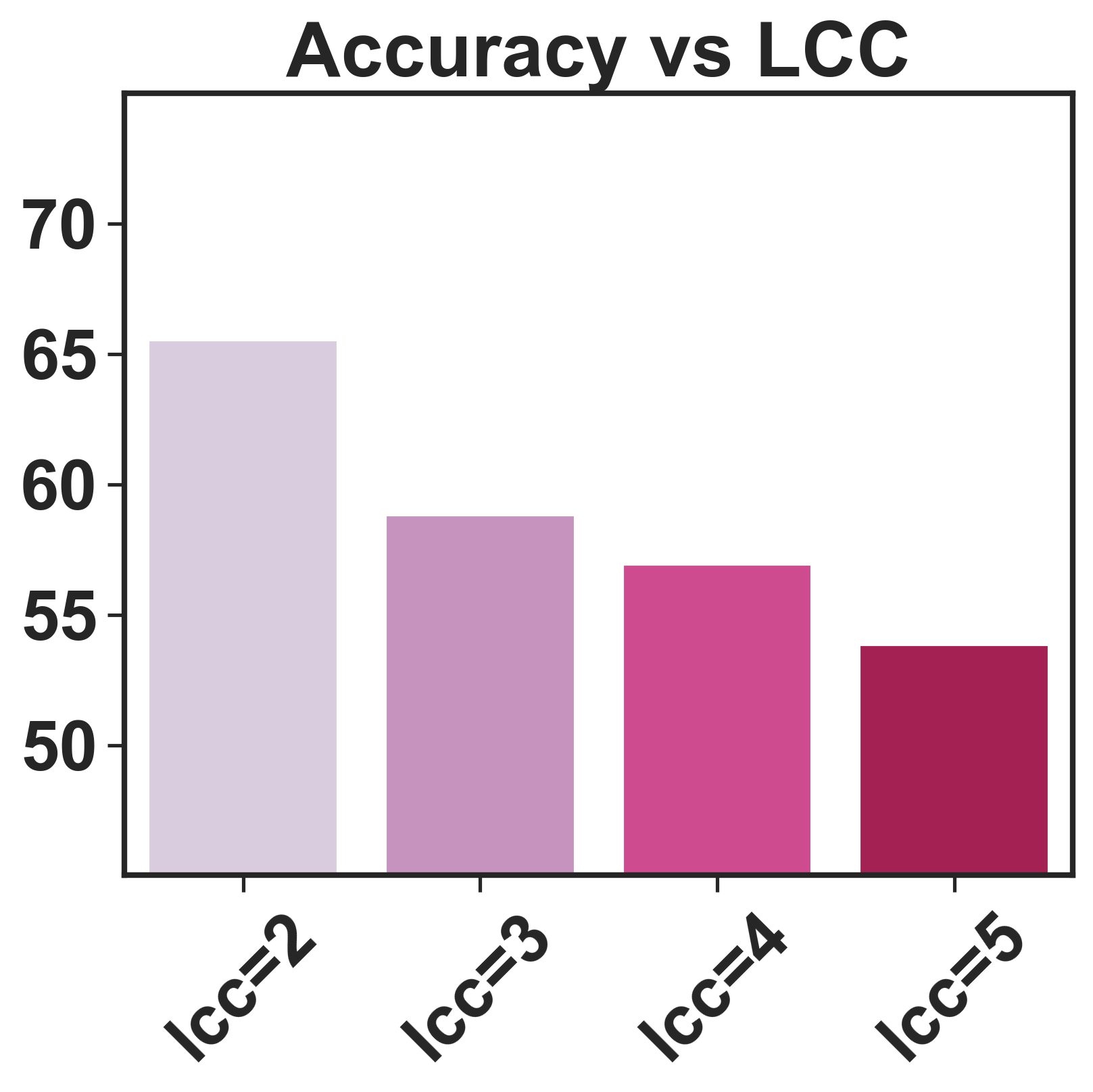}
  \end{minipage}\hfill 
  \begin{minipage}{0.4\columnwidth} 
    \centering
    \includegraphics[height=2.6cm]{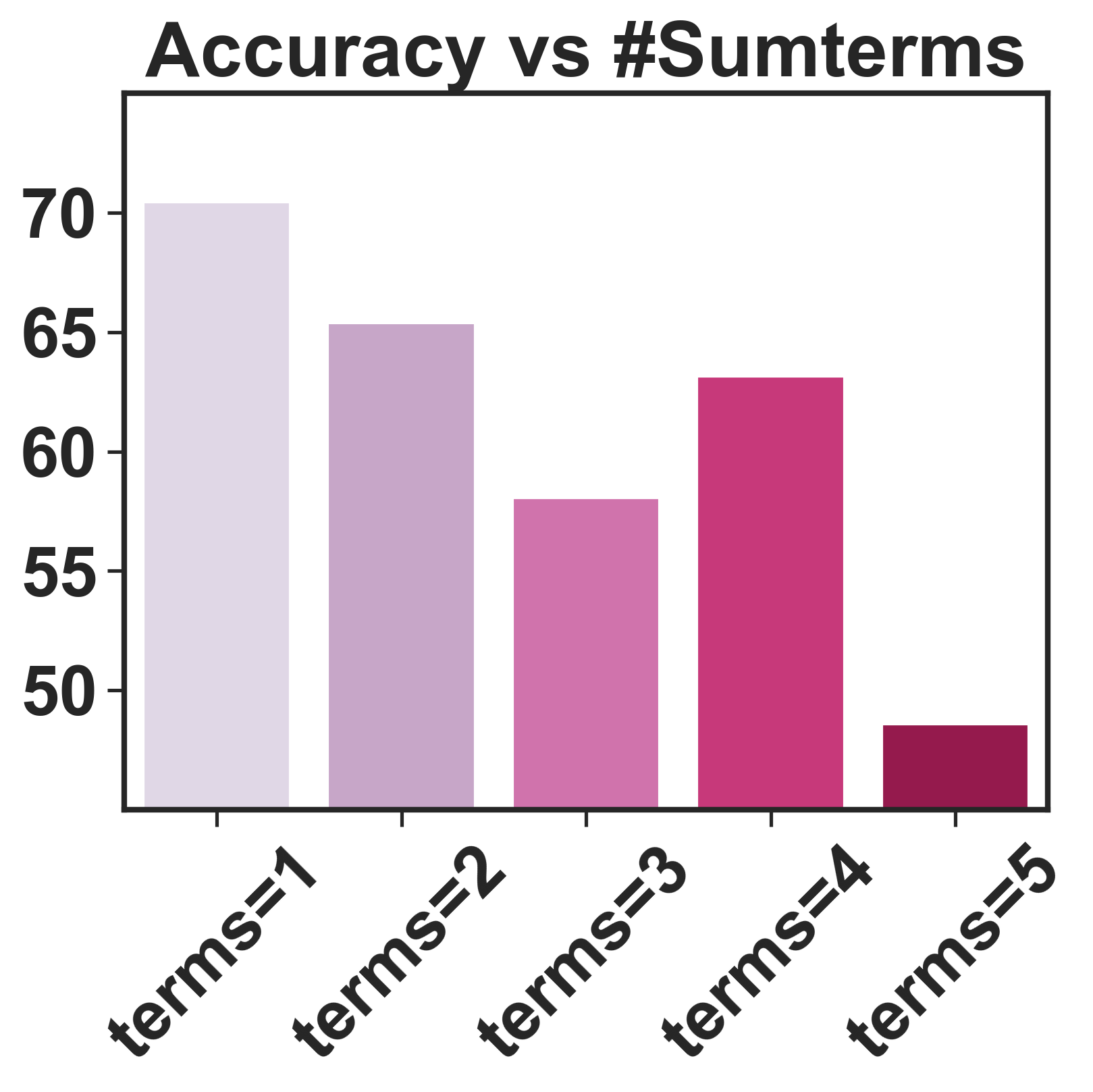}
  \end{minipage}\hfill
  \begin{minipage}{0.25\columnwidth} 
    \centering
    \includegraphics[height=2.6cm]{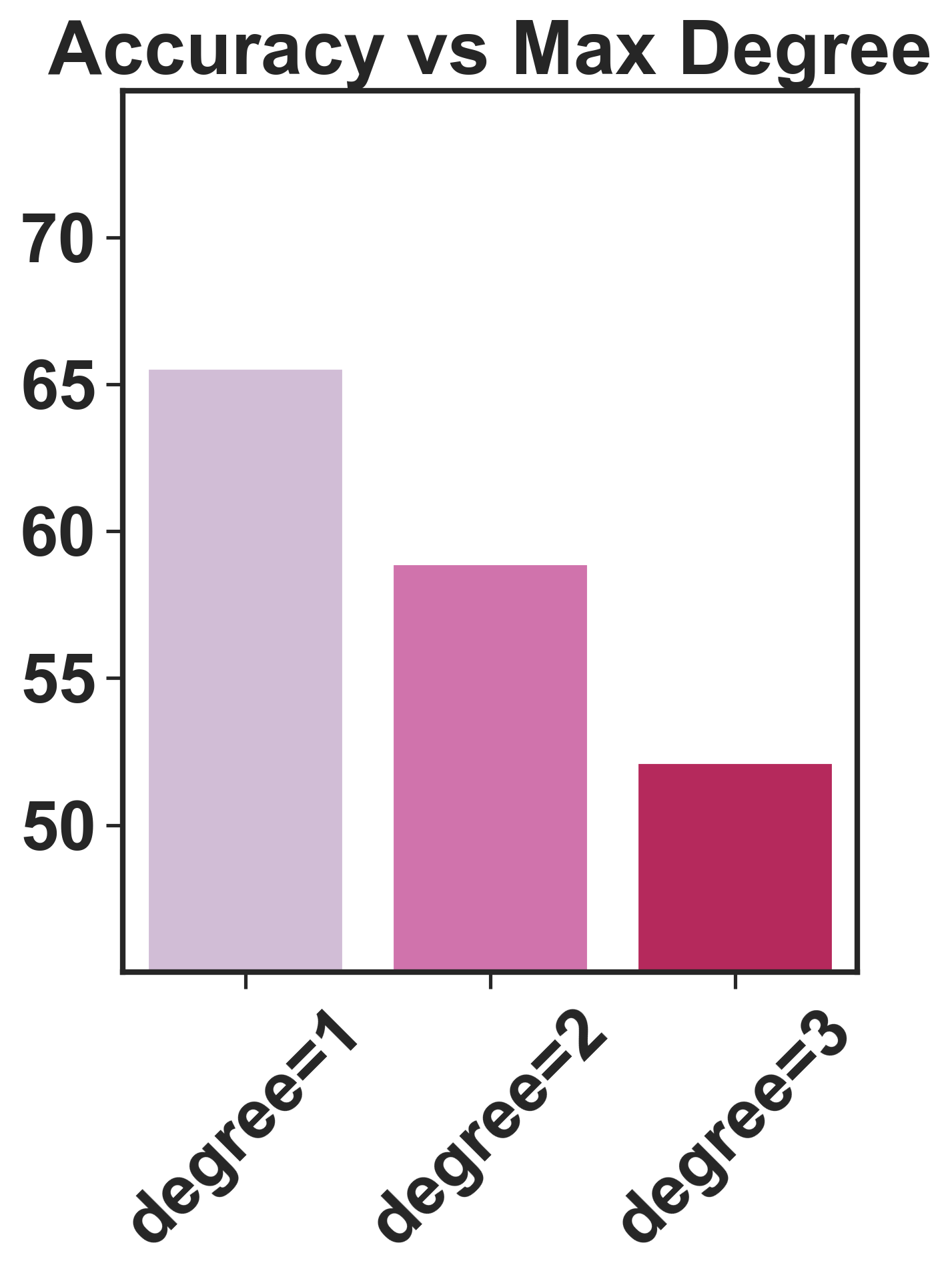}
  \end{minipage}
  
  \caption{Increasing complexity in evaluation samples.}
  \label{fig:complexity}
\end{figure}

\begin{figure*}[h] 
    \centering
    \includegraphics[height=3.4cm]{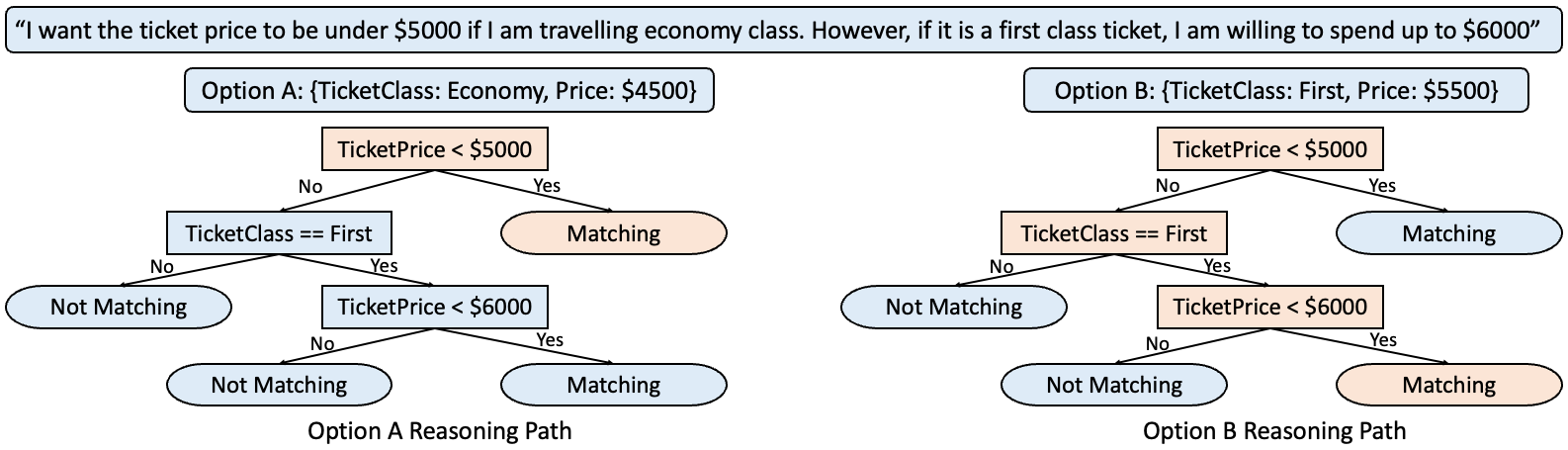}
    \caption{Sample user requirement and two hypothetical flight options.}
    \label{fig:entropy_example}
\end{figure*}

Given a fixed schema for the flight options, the number of sum terms in the POS expression as well as the LCC in the dependency graph are indicative of the \textit{reasoning width} and, in turn, the compositional complexity of the user criteria. The LCC is the length of the largest chain of slots - the possible values of which are dependent on one another through OR conditions (represented by edges in the dependency graph). This metric effectively reflects the breadth of parallel computation or \textit{reasoning width} required to accurately infer the given user criteria. Since increased branching in the dependency graph suggests a greater conditional complexity in user criteria, we also analyze model performance with increasing \textit{maximum degree} of the dependency graph. This gives us the extent of conditioning on a single slot value. In Figure~\ref{fig:complexity} we observe the decline in model performance with increased complexity as indicated by these factors. 

\subsection{Quantifying Confusion in Answer Choices through Entropy}

Numerous recent studies have explored how deep learning models, specifically transformer-based architectures, achieve success by exploiting shortcuts \citep{geirhos2020shortcut, liu2022transformers, lazy, 10.1145/3596490} and relying on spurious correlations present in the training data \citep{ZhangIJCAI23, PrOntoQA, PrOntoQAOOD}. Recently, \citet{dziri2023faith} utilized relative information gain of individual output elements in partially correct answers to explain surface pattern understanding in LLMs. In the same vein, we employ entropy as a metric to measure the confusion that might be caused due to conditions in the user query for a given flight option. We do this in an attempt to demystify how language models may succeed at some and fail at other queries with similar levels of complexity. To illustrate this, we take an example user requirement, and two hypothetical and simplified flight options as shown in Figure~\ref{fig:entropy_example}. Additionally, we show the reasoning path that must be navigated in each case for a successful outcome.

\begin{table}[H]
    \small
    \centering
    \begin{tabular}{lcc}
    \toprule
          & Option A & Option B\\ 
    \midrule
         \textsf{Price} < \$5000 & 1 & 0\\ 
    \midrule
         \textsf{TicketClass} = Economy & 1 & 0\\ 
    \midrule
         \textsf{Price} < \$6000 & 1 & 1\\ 
    \midrule
         \textsf{TicketClass} = First & 0 & 1\\ 
    \midrule
         $p_{sat}$ & 0.75 & 0.5\\
    \midrule
         $p_{\bar{sat}}$ & 0.25 & 0.5\\ 
    \midrule
         Entropy & 0.81125 & 1.0\\ 
    \bottomrule
    \end{tabular}
    \caption{Satisfaction of primitives and entropy.}
    \label{tab:entropy_example}
\end{table}

We observe how option B in our example leads to a more convoluted reasoning path, whereas the model is able to bypass considerable conditional overhead in the case of Option A. For the purpose of quantifying this more generally, we observe the compositional primitives (values attached to individual slots in the POS expression) in each sample and attach a binary value indicating if the primitive is satisfied. For the example in Figure~\ref{fig:entropy_example}, we show the primitives and the corresponding values of both options in Table~\ref{tab:entropy_example}. We also show the probability of a primitive being satisfied($p_{sat}$) and being unsatisfied($p_{\bar{sat}}$) by the flight option under consideration, as well as the final entropy.

\begin{figure}[H]
    \centering
    \includegraphics[height=3.4cm]{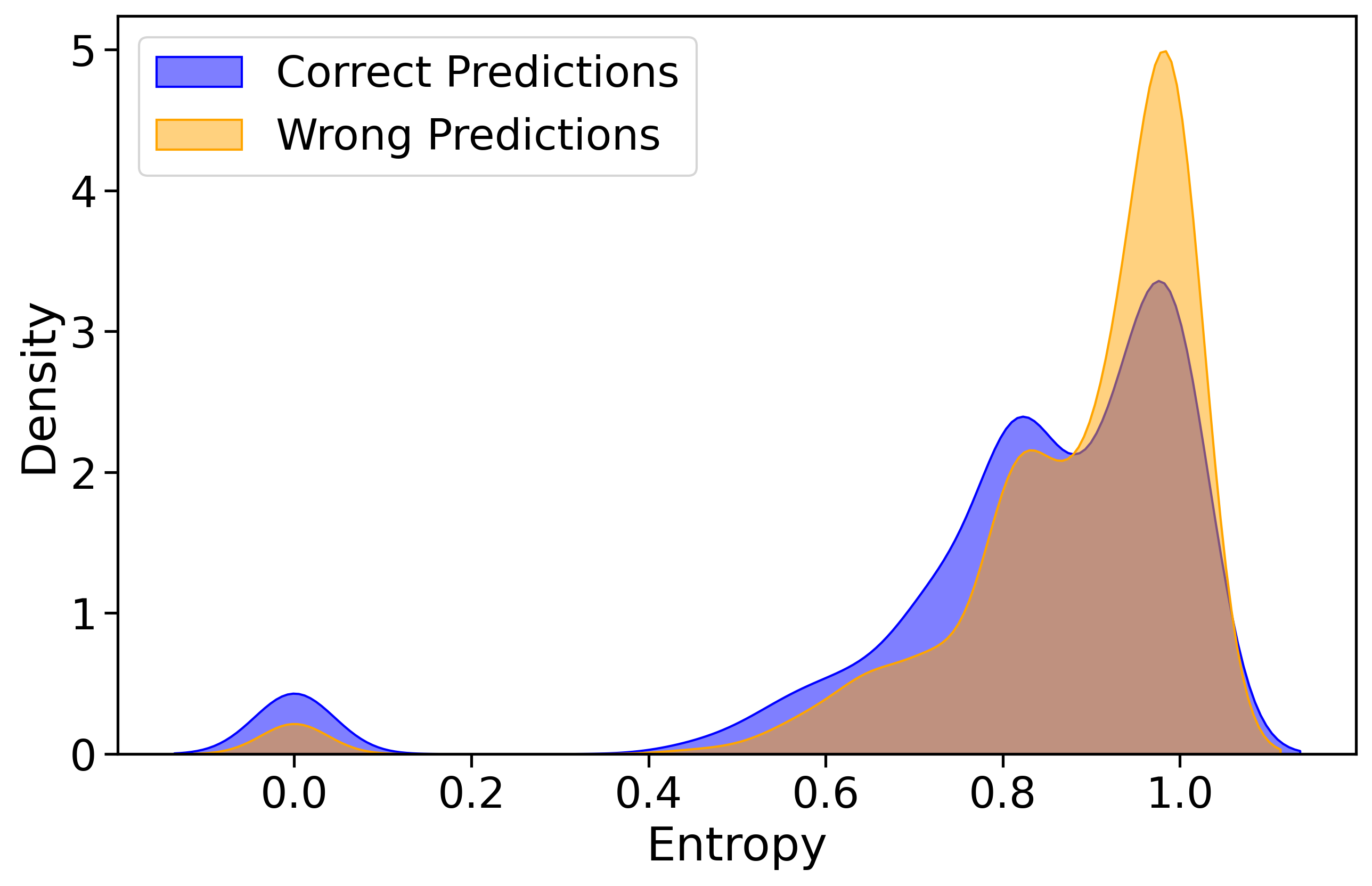}
    \caption{Effect of increasing entropy in answer choices.}
    \label{fig:entropy_violin}
\end{figure}

Entropy due to user criteria for each option can then be computed using the formula in Equation~\ref{eq:entropy}. Higher uncertainty leads to greater entropy in Option B as opposed to Option A, indicating a greater conditional overhead.
\begin{equation}
    H(X)=-(p_{sat}logp_{sat}+p_{\bar{sat}}logp_{\bar{sat}})
  \label{eq:entropy}
\end{equation}
In our analysis, we take the entropy values of the correct answer choice for each sample. Figure \ref{fig:entropy_violin} shows the densities of entropy values for the correct and wrong predictions of GPT-4 Turbo. While correct predictions exceed wrong predictions at lower entropy values, an abrupt surge in wrong predictions is observed at higher entropy levels. Thus, entropy gives us yet another measure of conditional complexity from the perspective of the answer choices rather than just the query, and helps explain why a model might exhibit inconsistent results across user queries of similar complexity.

\subsection{Robustness to Unconventional User Needs}
\label{sec:unconventional}

Several contemporary studies have sought to examine the robustness of language models by studying their resilience to out-of-distribution data \citep{koh2021wilds, wang2023robustness} or through adversarial attacks and input perturbations \citep{gardner2020evaluating, goel2021robustness, subhash2023universal, sanyal-etal-2022-robustlr, yuan2023can}. In our work, we challenge models through atypical user requirements in order to assess bias from pretraining and robustness to unorthodox and nontraditional queries. We segregate queries into "Regular" and "Atypical" groups as described in Section~\ref{sec:primitive_generation}. In Table~\ref{tab:results_main}, we contrast model performance on samples that describe such unconventional user needs versus those that do not. While most models in our testing show a decay in performance, the impact is more noticeable on better performing models such as GPT-4 Turbo. The in-context example used for all queries when testing with CoT includes two such primitives (ticket price $>$ 1800, carbon emission above average). We observe that the decline in performance is less pronounced with CoT.

\section{Related Work}

\noindent \textbf {Reasoning Challenges in NLP.} Our work extends the existing line of research on evaluating natural language processing (NLP) systems on different facets of reasoning - most notably commonsense question-answering \citep{talmor-etal-2019-commonsenseqa, huang-etal-2019-cosmos}, physical reasoning \citep{bisk2020piqa}, social interaction \citep{sap-etal-2019-social}, mathematical reasoning \citep{cobbe2021gsm8k, amini-etal-2019-mathqa, miao-etal-2020-diverse, hendrycksmath2021}, story completion \citep{zellers2019hellaswag}, temporal reasoning \citep{zhou2019going, tan-etal-2023-towards} abductive reasoning \citep{Bhagavatula2020Abductive} and pronoun resolution \citep{10.1145/3474381}. Different from these benchmarks, \dataname\ introduces a unique and substantial challenge for LLMs in the form of conditional and compositional reasoning.

Among these, ConditionalQA \cite{sun-etal-2022-conditionalqa} is arguably the most comparable to \dataname\ in terms of the skills it assesses. While ConditionalQA focuses on the reading comprehension of conditionally-complex policy documents, \dataname\ further tests the alignment/grounding ability of language models as (conditionally complex) user preferences have to be matched with multiple (5) flight schemas. The user requirements in \dataname\ are deliberately generated to introduce conditional complexity through our pipeline. These factors result in a greater number of reasoning paths (reasoning width) and the number of variables the model has to simultaneously consider when answering a question. Our method provides a controllable way to adjust for this complexity by a simple adjustment of parameters such as slots/minterms as described in Section \ref{subsec:pos_gen}. Thus, \dataname\ can be scaled to more complex examples in the future and also adapted to different domains.


\noindent \textbf {Benchmarks on Propositional Logic.} \dataname\ also aligns with the considerable body of work on evaluating logical reasoning in language models. The RuleTaker \citep{clark2021transformers} and ProofWriter \citep{tafjord2021proofwriter} datasets proposed a modern approach to evaluating logical reasoning through a task involving assignment of binary labels to candidate implications following a set of premises expressed in natural language. The datasets emulate a \textit{linear} deductive chain of reasoning of varying depths given a set of facts and rules, with ProofWriter augmenting this task through intermediate conclusions and proof generation. LogicNLI \citep{tian-etal-2021-diagnosing} provides a more comprehensive diagnostic benchmark involving reasoning through all seven fundamental logics (conjunction, disjunction, negation, implication, equation, universal and existential quantifiers). It contains an additional "paradox" label implying a situation where both the hypothesis as well as its negative proposition can be simultaneously entailed to the premise through different reasoning paths. This facilitates a non-linear reasoning, but is still limited to two contradictory reasoning paths. The FOLIO \citep{han2022folio} dataset boasts a higher vocabulary size due to a hybrid annotation approach but again consists of linear reasoning chains. Along similar lines, ProntoQA \citep{saparov2022language} proposes a first-order logic benchmark using a linear ontology which might be fictional. This is done to prevent LLMs from predicting correct outcomes through spurious correlations in their pretraining corpus. 

The benchmarks described here are primarily focused on the evaluation of deductive reasoning. In contrast, \dataname\ offers a more realistic grounding task with an emphasis on if-then reasoning which leads to many candidate reasoning paths for each answer choice. While deductive reasoning may involve a broader range of logical structures, conditional reasoning is a subset which deals specifically with the relationships and implications of conditional statements. Our dataset consists of a large vocabulary size and context length per sample, leading to greater linguistic diversity, and a higher reasoning width than other benchmarks in logical reasoning. Questions are designed to test for robustness against rare and unconventional user requirements and bring to the fore model bias from pretraining data. Also, unlike most other benchmarks, we do not attempt to evaluate logical reasoning in isolation - our task might require abilities such as temporal or mathematical reasoning.


\noindent \textbf {Compositional Generalization.} Samples in \dataname\ consist of novel combination of primitives expressed as user requirements in a flight-booking task. Such reasoning falls under the umbrella of compositional generalization - an area that has garnered increasing interest recently. \citet{hosseini-etal-2022-compositional} highlight the relative generalization gap with in-context learning between in-distribution and out-of-distribution samples in various semantic parsing tasks. \citet{dziri2023faith} demonstrates how transformer-based LLMs may solve compositional tasks by reducing them to linearized subgraph matching. By establishing a computational graph for each problem, the authors are able to define computational complexity by metrics such as the reasoning depth and width which correspond to levels in multi-hop reasoning and average parallelism respectively. Unsurprisingly, increased task complexity leads to a rapid decay in model performance under various settings.

Our findings largely concur with previous literature on compositional reasoning. However, results on \dataname\ reveal that even the most advanced LLMs struggle at relatively low levels of compositional complexity when juxtaposed with conditional reasoning and grounding. While \citet{dziri2023faith} demonstrated their results using problems such as multi-digit multiplication, dynamic programming, and Einstein's puzzle - we release a new dataset that is anchored on a practical, real world use-case of parsing complex user criteria and grounding to a fixed schema representing a flight option. \dataname\ contains a high semantic coverage and we posit that it would be of interest to the NLP community as a hard evaluation set to benchmark compositional generalization in LLMs.


\noindent \textbf {Dialogue-State Tracking.} Finally, while our task is reminiscent of a single turn in a dialogue state tracking system, it goes one step further to test a language model's grounding ability to match a flight schema with the user query. Most schema-guided dialogue datasets \citep{rastogi2020towards, lee2022sgd} consist of fixed slot values and filtering of available options is handled through external systems (e.g. api's). Slot values in \dataname\ are fuzzy due to conditional constraints on the primitives - in Figure~\ref{fig:entropy_example}, \textsf{TicketPrice} may take on different values based on \textsf{TicketClass}. \dataname\ consists of examples with varying levels of compositional complexity due to long and complex user requirements. This differentiates it from the majority of schema-guided dialogue datasets where the primary objective is goal identification and tagging of slot values. These tasks, while challenging in their own respect, do not engage a models' compositional reasoning ability to the same extent.

\section{Conclusion}

Modern LLMs have demonstrated remarkable advancements in many tasks including those that are inherently compositional and necessitate conditional reasoning such as mathematical problem solving, and code generation and interpretation. However, discerning genuine reasoning from mere rote learning and shallow understanding continues to be a focal point of study. Though proficient at answering questions of seemingly greater complexity, we show that they can struggle on the same skills when presented with an unfamiliar task setting. While problem size does have an impact, even the less complex samples in our dataset are challenging to the best language models today.

Beyond introducing a new benchmark dataset, we conduct a thorough analysis of the effects of increasing complexity, including advanced prompting techniques, and robustness to atypical queries. Our results uncover a substantial disparity in the performance of competing language models, a distinction that is not as pronounced in most other evaluation benchmarks and highlights their respective abilities in tackling novel challenges. Our data generation process is largely automatic, with human validation at the last step. In addition to the dataset and the evaluation script, we release code for the data generation which can be easily extended to generate more examples, and increase diversity (through different slots) as well as complexity. With minor modifications, the task can be further complicated by incorporating queries with multiple answers and questions that require other forms of logical reasoning such as aggregation (e.g., "Give me the cheapest flight matching my criteria?") and existential quantification (e.g., "Is there a first class seat under \$5000?"), greater world knowledge (e.g., "I'd like to avoid layovers in Europe") etc., which we leave for future work.

\section{Limitations}

\dataname\ consists entirely of samples in the flight-booking domain. This scenario is popular and widely used in training and evaluation benchmarks for dialogue state tracking, planning etc. Due to the general nature of the primitives used in our flight requirements, we are confident that the results and insights would be applicable to a wide array of domains. However, this has not been empirically validated and we leave the extension of \dataname\ to other domains as a topic for future research.

To isolate unconventional user requirements, we identify primitives that are uncommon in typical flight reservation scenarios (e.g., "I want more than two layovers"). However, the criteria for segregation involves a degree of subjectivity. Furthermore, conventional primitives can be combined in unconventional ways using conditional formats (e.g., "If the flight is after 7 pm, I want the carbon emissions to be below average"), which our approach for identifying unconventional requirements does not account for. Consequently, further investigation is needed to evaluate model robustness to unconventional requirements that significantly deviate from patterns likely encountered in training data.


Finally, we assess the performance of LLMs using both CoT and L2M prompting techniques. However, L2M requires several decomposition steps, resulting in multiple prompts to the various LLMs for each test sample. Given the high inference cost associated with this approach, our evaluation is limited to a subset of 200 samples. While the results suggest that \dataname\ remains a challenging benchmark even with L2M prompting, they do not offer a full assessment of individual LLM performance under this setting.

\bibliography{custom}

\onecolumn
\appendix

\clearpage

\section{Sample Flight Schema}
\label{sec:schema}

\begin{figure}[H]
    \centering
    \includegraphics[width=\textwidth]{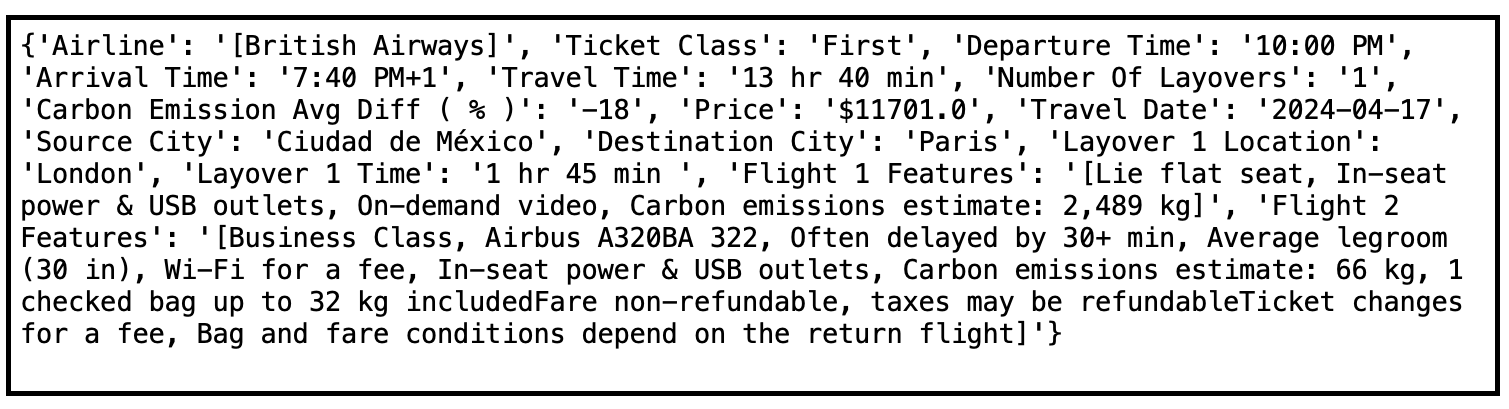}
    \caption{Schema for British Airways flight between Mexico City and Paris on 04/24/2024}
    \label{fig:flight_schema}
\end{figure}

\section{Samples from different slot/minterm configurations}
\label{sec:example_generations}

\subsection{2 slots, 2 minterms}

\begin{figure}[H]
    \centering
    \includegraphics[width=\textwidth]{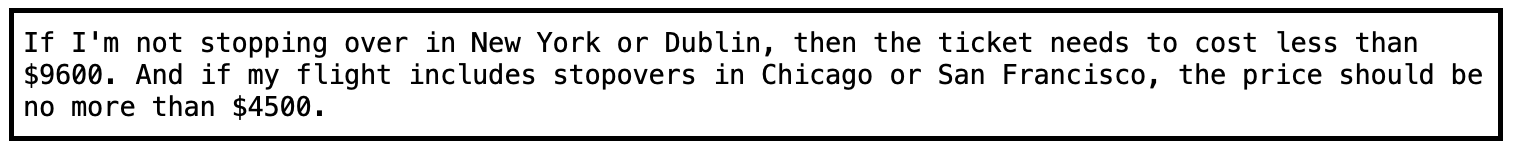}
    \caption{Sample query in \dataname\ with 2 slots and 2 minterms.}
\end{figure}

\subsection{3 slots, 2 minterms}

\begin{figure}[H]
    \centering
    \includegraphics[width=\textwidth]{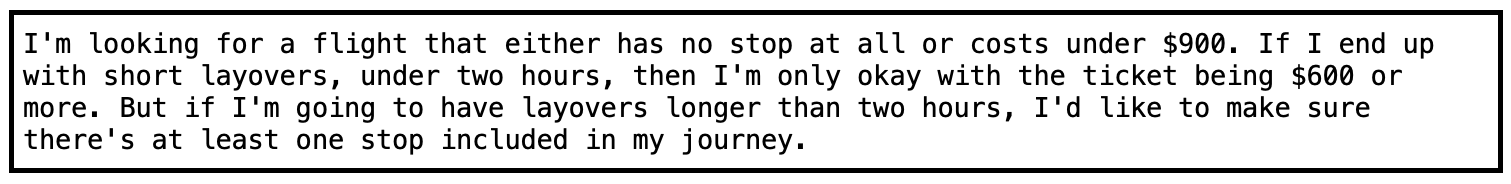}
    \caption{Sample query in \dataname\ with 3 slots and 2 minterms.}
\end{figure}

\subsection{4 slots, 2 minterms}

\begin{figure}[H]
    \centering
    \includegraphics[width=\textwidth]{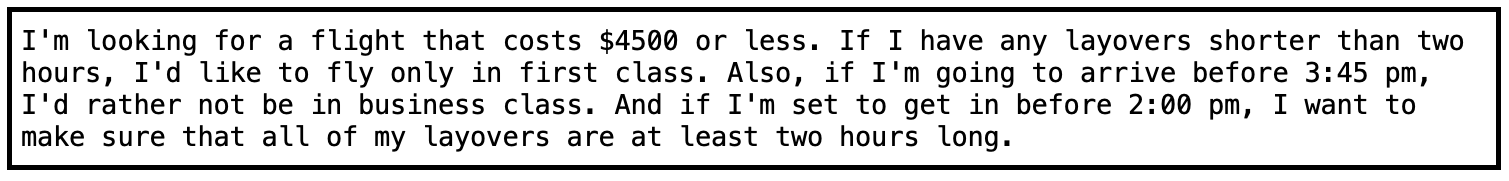}
    \caption{Sample query in \dataname\ with 4 slots and 2 minterms.}
\end{figure}

\subsection{4 slots, 3 minterms}

\begin{figure}[H]
    \centering
    \includegraphics[width=\textwidth]{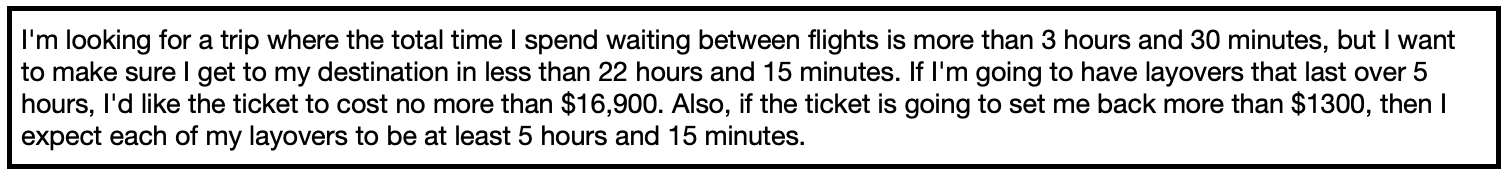}
    \caption{Sample query in \dataname\ with 4 slots and 3 minterms.}
\end{figure}

\subsection{5 slots, 2 minterms}

\begin{figure}[H]
    \centering
    \includegraphics[width=\textwidth]{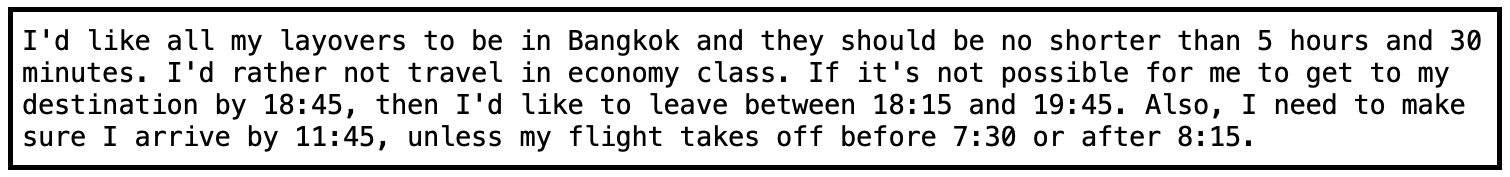}
    \caption{Sample query in \dataname\ with 5 slots and 2 minterms.}
\end{figure}

\subsection{6 slots, 2 minterms}

\begin{figure}[H]
    \centering
    \includegraphics[width=\textwidth]{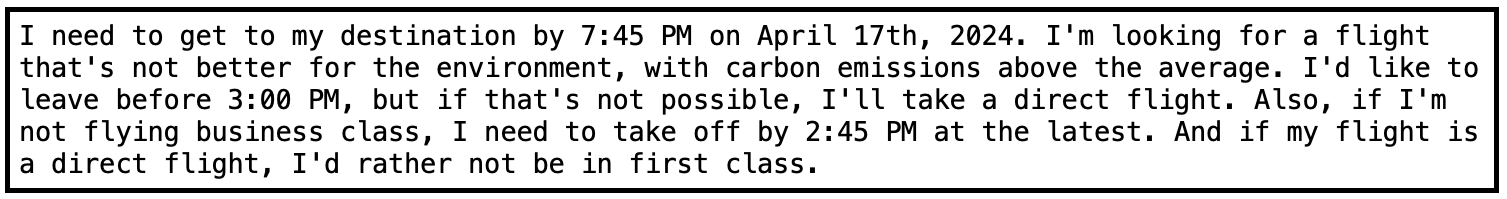}
    \caption{Sample query in \dataname\ with 6 slots and 2 minterms.}
\end{figure}

\section{Prompts used in Model Evaluation}
\label{sec:testing_prompts}

\begin{figure}[h]
    \centering
    \includegraphics[width=\textwidth]{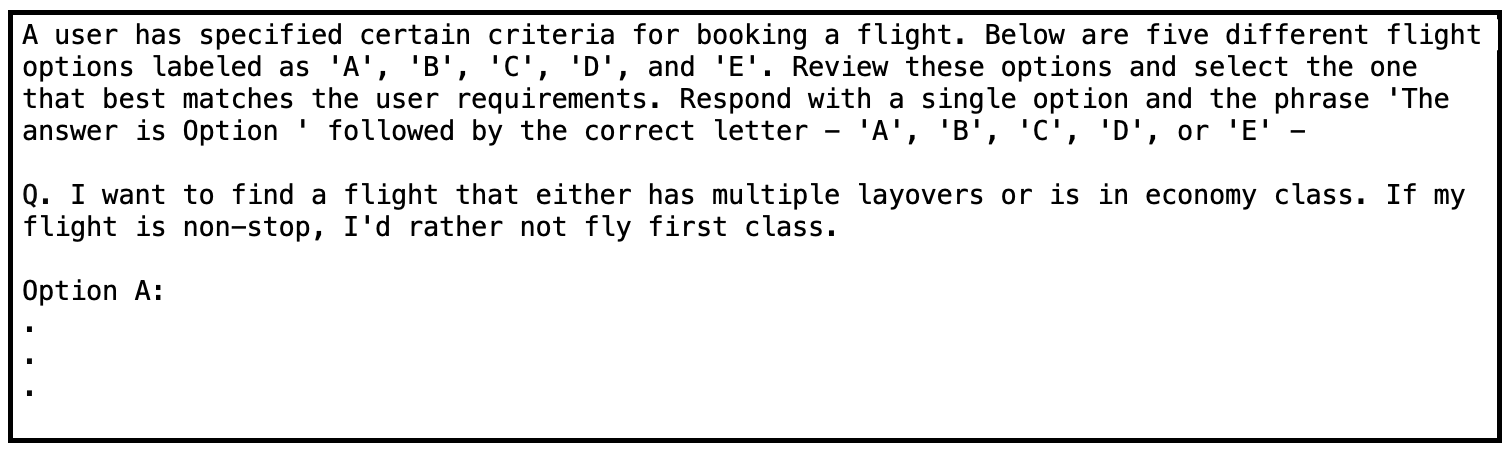}
    \caption{Evaluation prompt without CoT.}
    \label{fig:question_prompt}
\end{figure}

\begin{figure}[h]
    \centering
    \includegraphics[height=\textheight - 1cm]{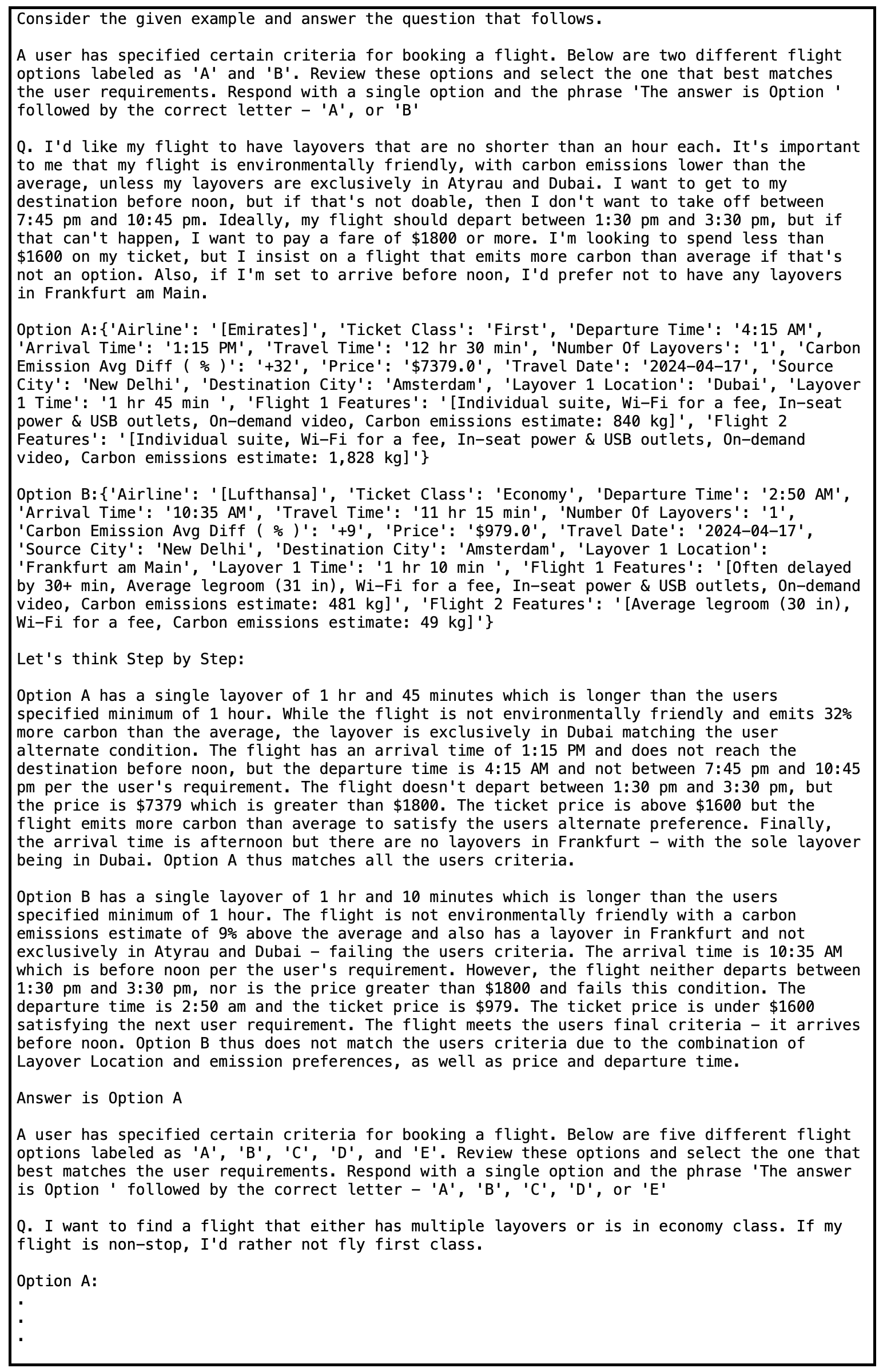}

    \caption{Evaluation prompt using CoT-partial.}
    \label{fig:cot_partial}
\end{figure}

\section{End-to-End Generation Process}
\label{sec:llm_paraphrase}

\begin{figure}[H]
    \centering
    \includegraphics[width=\textwidth]{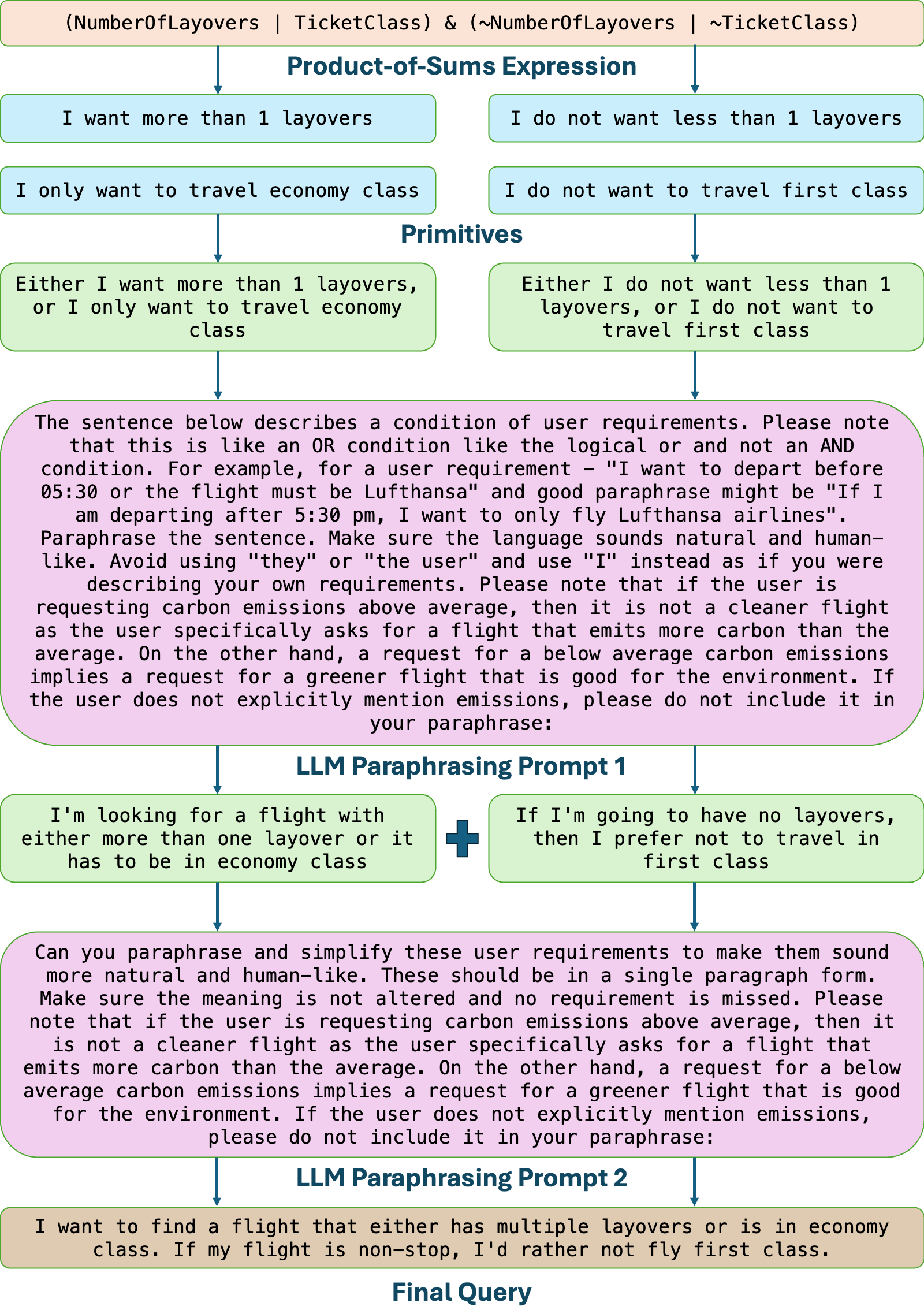}
    \caption{End-to-End query generation with 2 slots and 2 minterms.}
    \label{fig:llm_paraphrase}
\end{figure}

\end{document}